\pdfoutput=1

\documentclass[11pt]{article}

\usepackage[final]{acl}

\usepackage{times}
\usepackage{latexsym}
\usepackage{booktabs}
\usepackage{multirow}

\usepackage[T1]{fontenc}
\usepackage{kotex}
\usepackage{tabularx}

\usepackage[utf8]{inputenc}

\usepackage{microtype}
\usepackage{amsmath} 
\usepackage{inconsolata}

\usepackage{graphicx}

\usepackage{placeins}

\title{Controlling Language Confusion in Multilingual LLMs}

\author{
Nahyun Lee{\textsuperscript{1,3}} \quad 
Yeongseo Woo{\textsuperscript{1}} \quad 
Hyunwoo Ko{\textsuperscript{2,3}} \quad 
Guijin Son{\textsuperscript{2,3}}  
 \\ \\
Chungang University{\textsuperscript{1}} \quad 
OneLineAI{\textsuperscript{2}}  \quad MODULABS{\textsuperscript{3}}
\\
\texttt{naa012@cau.ac.kr} \quad 
\texttt{spthsrbwls123@yonsei.ac.kr}\\ 
}

\begin{document}
\maketitle
\begin{abstract}


Large language models often suffer from language confusion, a phenomenon in which responses are partially or entirely generated in unintended languages. This critically degrades the user experience, especially in low-resource settings. 
We hypothesize that this issue stems from limitations in conventional fine-tuning objectives, such as supervised learning, which optimize the likelihood of correct tokens without explicitly penalizing undesired outputs such as cross-lingual mixing. Analysis of loss trajectories during pretraining further reveals that models fail to distinguish between monolingual and language-mixed texts, highlighting the absence of inherent pressure to avoid such confusion. In this work, we apply ORPO, which adds penalties for unwanted output styles to standard SFT, effectively suppressing language-confused generations. ORPO maintains strong language consistency, even under high decoding temperatures, while preserving general QA performance. 
Our findings suggest that incorporating appropriate penalty terms can effectively mitigate language confusion in multilingual models, particularly in low-resource scenarios.

\end{abstract}

\section{Introduction}

Scaling large language models has empirically delivered substantial gains in multilingual capabilities~\citep{hurst2024gpt, cohere2025command, yang2025qwen3}, across diverse tasks such as machine translation~\citep{alves2024tower}, summarization~\citep{forde-etal-2024-evaluating}, and reasoning~\citep{son2025linguistic}. 
However, despite their growing capabilities, LLMs often suffer from language confusion~\citep{marchisio2024understanding}, a failure mode in which outputs inadvertently blend multiple languages. This hampers real-world deployment of LLM systems as even the most minor language confusion may be critical to user experience~\citep{son2024llm}. This issue is particularly pronounced in low-resource settings, where limited supervision exacerbates cross-lingual interference~\citep{ arivazhagan2019multilingual, wang2023seaeval}.

However, little research has been conducted on \emph{why} such behavior may happen. In this work, we draw inspiration from the training methodology proposed by \citet{hong2024orpo}, which applies supervised fine-tuning to preferred generation styles while imposing penalties on disfavored ones.

In this work, we conduct two experiments to investigate whether language confusion arises from the absence of an explicit penalty against undesired languages.

First, we track the training loss of two model families (SmolLM2~\citep{allal2025smollm2} and OLMo2~\citep{olmo20242}) throughout their pretraining process. 
In both cases, the loss of language‐confused outputs steadily decreases over time, indicating that the models do not learn to disfavor confused generations. 
Additionally, by using ORPO~\citep{hong2024orpo} for an additional three epochs of fine-tuning, we show that introducing an explicit penalty against unwanted languages effectively restricts language confusion.

\section{Preliminaries}

\subsection{Related Works}

\paragraph{What is language confusion?} Language confusion, also known as language mixing or code-mixing, occurs when two or more languages are mixed within a single utterance~\citep{chen2024confusion, yoo2024code}. This phenomenon is particularly prevalent in low-resource languages~\citep{arivazhagan2019multilingual} and even appears in state-of-the-art models~\citep{VictorRM_2025_o3}. Diverse discussions have emerged regarding language confusion. Although it can sometimes support multilingual transfer~\citep{wang2025investigating}, mixed-language responses may undermine user experience, as they can be perceived as signs of incompetence~\citep{son2024llm}.

\subsection{Quantifying Language Confusion}

Measurement of language confusion can be challenging, as LLM judges~\citep{zheng2023judging} remain unreliable~\citep{son2024mm}, and rule-based methods cannot distinguish genuine confusion from legitimate uses of foreign language (e.g., abbreviations). In this work, we leverage two metrics Word Precision Rate (WPR) and Language Precision Rate (LPR) proposed by \citet{marchisio2024understanding}.

\textbf{WPR} computes the overall fraction of tokens produced in the target language, offering a granular view of how consistently a model sticks to one language. Where  \(\mathcal{T} = \bigcup_{i=1}^{N} T_i\) is the set of all valid tokens across \(N\) outputs, WPR is defined as:

\begin{equation}
\frac{\bigl|\{\,t \in \mathcal{T} : \text{is\_Korean}(t)\}\bigr|}{|\mathcal{T}|}
\end{equation}

\textbf{LPR} counts the proportion of sentences in which at least 90\% of tokens belong to the target language, thereby penalizing any cross‐lingual intrusions. Where \(I(\cdot)\) denotes the indicator function and \(s_i\) the \(i\)-th sentence, LPR is defined as:

\begin{equation}
\frac{1}{N}\sum_{i=1}^{N} 
I\!\Bigl(
\frac{\bigl|\{\,t \in s_i : \text{is\_Korean}(t)\}\bigr|}
{\bigl|\{\,t \in s_i : \text{is\_valid}(t)\}\bigr|}
\;\ge\;0.9
\Bigr)
\end{equation}

Additionally, as noted above, rule-based metrics alone cannot distinguish true language confusion from minor lexical variations, such as numerals, named entities, or common loanwords. Therefore, alongside WPR and LPR, we also report the proportion of responses with WPR and LPR exceeding 0.9. Empirically, we observe that many such responses remain perfectly acceptable sentences containing a few legitimate English terms. For examples of sentences with varying WPR and LPR levels, see Appendix~\ref{appendix:different_levels_of_wpr_lpr}.


\section{Experimental Setup}

\subsection{Dataset Preparation}

To facilitate pairwise preference learning, we constructed instruction-centered triplet datasets. 
Each triplet comprises a Korean prompt \textit{(input)}, a fully Korean response \textit{(chosen)}, and an alternative response exhibiting code-mixing or a full unexpected language \textit{(rejected)}.



\begin{figure}[h]
    \centering
    \includegraphics[width=1.0\columnwidth]{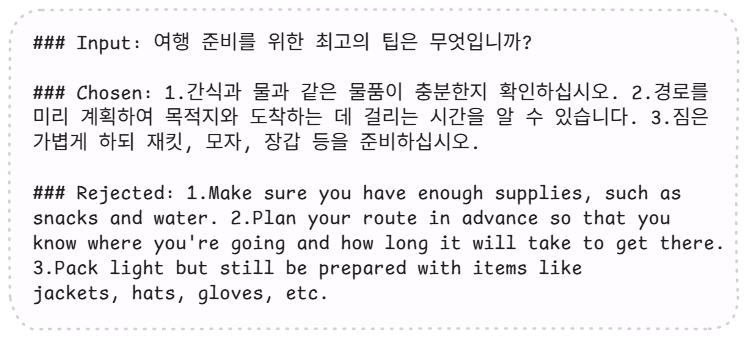}
    \caption{Dataset structure (OIG, Chosen-Rejected pair)}
    \label{fig:chip2data}
\end{figure}


We constructed three multilingual datasets based on existing Korean corpora, each designed to represent a different form of language confusion. 
The OIG dataset~\citep{oig2022, heegyu_oig_small_chip2_ko} and HC3 dataset~\citep{guo2023close, hc3ko} pair Korean prompts with rejected responses written entirely in English.
In contrast, the KoAlpaca dataset~\citep{koalpaca2023} introduces more nuanced confusion by synthetically injecting translated English or Chinese tokens into Korean outputs, resulting in code-mixed responses. 
Additional preprocessing and filtering steps are described in Appendix~\ref{appendix:dataset-preprocessing}.


\subsection{Experiment Setup}

We fine-tuned two publicly available instruction-tuned language models: SmolLM2-1.7B~\cite{allal2025smollm2} and OLMo2-7B~\cite{olmo20242}, selected for their ability to generate Korean text among lightweight open source models.
Detailed training configurations are provided in Appendix~\ref{appendix:orpo-training-config}.

\subsection{Evaluation Protocol}

We evaluate three model variants: \textbf{Base}, the original instruction-tuned model; \textbf{SFT}, supervised fine-tuned on Korean prompt–response pairs from the OIG dataset; and \textbf{ORPO}, fine-tuned using Odds Ratio Preference Optimization, on the same dataset.


\begin{table*}[h]
\centering
\small
\resizebox{\textwidth}{!}{%
\begin{tabular}{llcccccccccccc}
\toprule
& {\textbf{Model}} 
& \multicolumn{6}{c}{\textbf{SmolLM2-1.7B}} 
& \multicolumn{6}{c}{\textbf{OLMo2-7B}} \\

\cmidrule(lr){3-8} \cmidrule(lr){9-14}
& \textbf{Temperature} &\multicolumn{2}{c}{0.7} & \multicolumn{2}{c}{1.0} & \multicolumn{2}{c}{1.2} 
  & \multicolumn{2}{c}{0.7} & \multicolumn{2}{c}{1.0} & \multicolumn{2}{c}{1.2} \\
\cmidrule(lr){3-4} \cmidrule(lr){5-6} \cmidrule(lr){7-8}
\cmidrule(lr){9-10} \cmidrule(lr){11-12} \cmidrule(lr){13-14}
& & Base & ORPO & Base & ORPO & Base & ORPO & Base & ORPO & Base & ORPO & Base & ORPO \\
\midrule
\multirow{4}{*}{\textbf{Metric}} 
& WPR > 0.9 ratio   & 96.1\% & 100.0\% & 94.3\% & 100.0\% & 81.4\% & 100.0\% & 96.3\% & 99.8\% & 91.8\% & 99.9\% & 7.5\% & 99.0\% \\
& LPR > 0.9 ratio   & 92.6\% & 99.9\% & 88.5\% & 100.0\% & 71.2\% & 99.9\% & 71.2\% & 99.7\% & 46.0\% & 99.8\% & 0.5\% & 96.8\% \\
& Average WPR       & 0.9821 & 0.9999 & 0.9696 & 1.0 & 0.8953 & 0.9999 & 0.9818 & 0.9998 & 0.9576 & 0.9998 & 0.6799 & 0.9962 \\
& Average LPR       & 0.9681 & 0.9996 & 0.9496 & 1.0 & 0.8434 & 0.9999 & 0.9379 & 0.9992 & 0.8684 & 0.9995 & 0.3044 & 0.9881 \\
\bottomrule
\end{tabular}
}
\caption{Comparison of SmolLM2 and OLMo2 models across temperatures (Base vs. ORPO). All metrics are higher is better: higher values indicate stronger language consistency.}
\label{tab:main_result_wpr_lpr_table}
\end{table*}

\section{Main Results}
Prior work shows LLMs default to high-frequency, dominant-language tokens when uncertain, causing language confusion~\citep{marchisio2024understanding}. We hypothesize that the standard next-token prediction objective exacerbates this bias: softmax focuses probability mass on the correct token but does not explicitly penalize cross-lingual mixing.

\subsection{Loss-Based Diagnostic: Do LLMs Penalize Language Mixing?}

\begin{figure}[h]
    \centering
    \includegraphics[width=1.0\columnwidth]{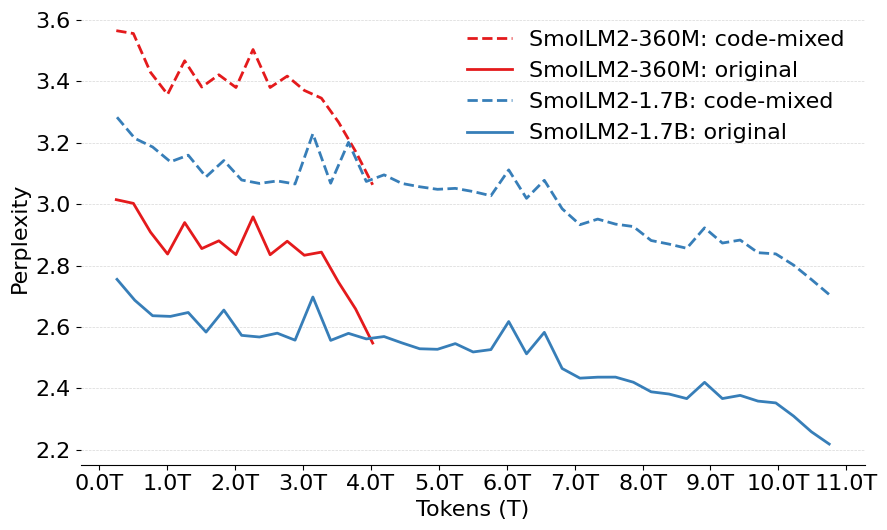}
    \caption{Average loss for monolingual and code-mixed responses across training tokens (SmolLM2)}
    \label{fig:loss_smollm}
\end{figure}

We begin with the observation that, during pretraining, neither SmolLM2~\citep{allal2025smollm2} model learns to penalize language confusion, as shown by their loss trajectories in Figure~\ref{fig:loss_smollm}.

In principle, a model that internalizes a robust linguistic preference should learn to assign lower loss to coherent Korean-only generations while preserving relatively higher loss for language-confused outputs. Contrary to expectations, we observe a monotonic decrease in loss for both chosen and rejected responses.
This trend may suggest that, in the absence of explicit preference signals, models eventually learn to prefer \emph{any} sequence of tokens they have seen during training, without distinguishing linguistically coherent and code-mixed outputs.
Such behavior persists up to the 7B scale, suggesting that model size alone cannot resolve the issue. See Appendix~\ref{appendix:loss-olmo} for results of OLMo2 models.

\subsection{Generation-level evaluation: WPR and LPR Comparison}

To evaluate the effectiveness of preference-based tuning method, we compare the generation performance of the Base and ORPO-tuned models using WPR and LPR under varying decoding temperatures.  
Each model generated responses for the same set of 1,000 prompts, repeated three times per prompt, and all reported scores are averaged across the three generations.  

As summarized in Table~\ref{tab:main_result_wpr_lpr_table}, we observe the following trends:

\begin{itemize}
    \item \textbf{ORPO-tuned models consistently outperform the Base models}, achieving near-perfect WPR and LPR even at high temperature settings (up to 1.2).

    \item \textbf{Temperature significantly impacts the Base models}. For instance, average LPR of the OLMo2 base model plummets to 0.3044 at a temperature of 1.2, indicating a severe degradation of linguistic consistency without preference-based fine-tuning.

\end{itemize}

\section{Additional Results}
\subsection{Comparison with other fine-tuning methods}

To evaluate how ORPO compares to other standard fine-tuning approaches, we conducted additional experiments using Supervised Fine-Tuning (SFT) and Direct Preference Optimization (DPO) under identical conditions.

Detailed results for both SmolLM2 and OLMo2 are presented in Appendix~\ref{appendix: WPR/LPR results across 3 levels of temperatures}.
Across both model families, ORPO consistently achieves high WPR and LPR scores, matching or slightly exceeding SFT and substantially outperforming DPO.

\subsection{Do fine-tuned models internalize penalties?}

\begin{figure}[h]
    \centering
    \includegraphics[width=1.0\columnwidth]{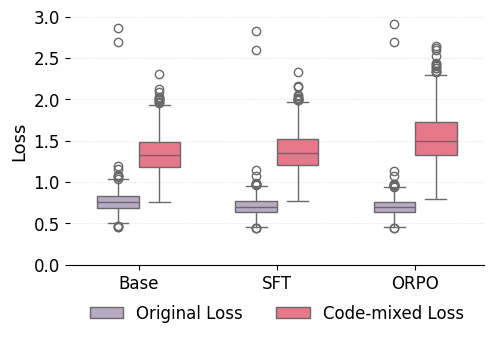}
    \caption{Loss of SmolLM2 models across tuning methods for both original and code-mixed responses}
    \label{fig:hc3-loss-smollm}
\end{figure}

To further investigate whether preference-based learning offers additional internal modeling advantages, we conduct a loss-based diagnostic analysis on the evaluation subset HC3 and compare the loss between original \textit{(chosen)} and code-mixed \textit{(rejected)} responses.

\begin{figure}[h]
    \centering
    \includegraphics[width=1.0\columnwidth]{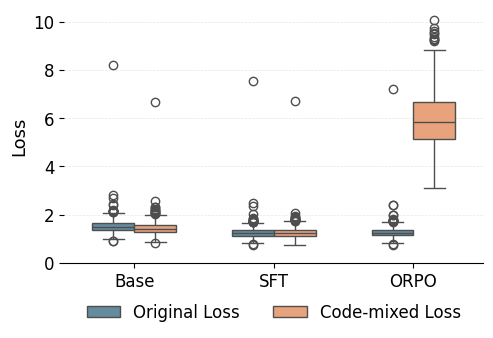}
    \caption{Loss of OLMo2 models across tuning methods for both original and code-mixed responses}
    \label{fig:hc3-loss-olmo}
\end{figure}

We found that ORPO assigns significantly higher loss to code-mixed responses compared to other models, indicating stronger penalization of language-confused outputs. 
On the HC3 evaluation set, ORPO yields an average delta loss of 0.8379 for SmolLM2 and 4.6778 for OLMo2-both the highest among all fine-tuning methods.
This increased separation suggests that ORPO fine-tuning more effectively reinforces internal preferences for linguistically consistent outputs, enabling more reliable discrimination between coherent and code-mixed generations (Figure~\ref{fig:hc3-loss-smollm} and \ref{fig:hc3-loss-olmo}).



\subsection{Does ORPO Fine-Tuning Lead to a Trade-off in General QA Capabilities?}

We assess whether ORPO fine-tuning, which mitigates language confusion, adversely affects general performance by evaluating our models on the HAE-RAE benchmark—a Korean multiple-choice QA suite covering general knowledge, history, loanwords, and rare vocabulary \citep{son2023hae}. 
We omit more challenging reasoning benchmarks due to the modest size of our models and limited training data. We compared three model variants: Base, SFT and ORPO fine-tuned model.

\begin{figure}[h]
    \centering
    \includegraphics[width=1.0\columnwidth]{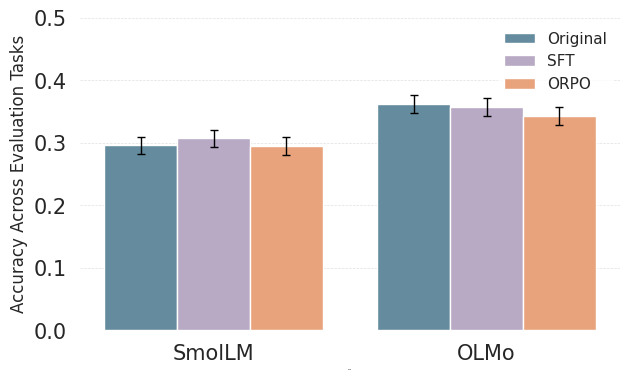}
    \caption{Average accuracy across training methods for SmolLM2 and OLMo2.}
    \label{fig:accuracy}
\end{figure}

Figure~\ref{fig:accuracy} reports the average accuracies in all subcategories for the SmolLM2 and OLMo2 models.
The results show no significant performance degradation in the three tuning methods.

These findings suggest that neither SFT nor ORPO introduces measurable harm to general QA capabilities.
In particular, ORPO maintains general QA performance while reducing language confusion.


\section{Conclusion}
This work investigates the underlying causes of language confusion in multilingual large language models and empirically demonstrates that penalizing undesired languages via preference optimization is an effective method for suppressing such behavior.


Our primary contribution is the demonstration that preference-based fine-tuning offers a highly effective solution. By fine-tuning models to prefer monolingual responses over language-confused ones, we achieve robust linguistic consistency without compromising general question-answering capabilities.

These results suggest that incorporating explicit preference signals during fine-tuning provides a promising approach for reinforcing linguistic fidelity in multilingual settings. Moreover, we suggest that future research may explore the use of penalty terms even in the pretraining phase to penalize language confusion earlier in the training effectively.

\section*{Limitations}

While our findings demonstrate the effectiveness of ORPO for mitigating language confusion, we acknowledge several limitations in this study.

First, our analysis does not include a sensitivity analysis of ORPO's hyperparameters. We used a fixed value ($\beta=0.1$) based on the original ORPO paper.  Future work should explore how varying this hyperparameter affects the trade-off between linguistic fidelity and general task performance.

Second, our experiments were conducted primarily on Korean-centric datasets and two specific model families (SmolLM2 and OLMo2).  Although the results are strong, further research is needed to ascertain whether our findings generalize to other languages and other model architectures.

Third, we did not perform an in-depth analysis of why ORPO consistently outperforms DPO. Further investigation is needed to fully understand the optimization dynamics behind this difference.

Finally, although we have detailed our experimental setup and dataset construction, we have not yet released the code and training artifacts. To facilitate reproducibility, we plan to make all code and training materials publicly available upon publication.

\section*{Acknowledgements}
This research was supported by Brian Impact Foundation, a non-profit organization dedicated to the advancement of science and technology for all.


\bibliography{custom}

\appendix

\section{Dataset preprocessing}
\label{appendix:dataset-preprocessing}

\textbf{KoAlpaca (Code-Mixed Rejection):}
We constructed this dataset using the KoAlpaca\footnote{\url{https://huggingface.co/datasets/beomi/KoAlpaca-v1.1a}} corpus, a Korean instruction-tuning dataset modeled after Stanford Alpaca~\citep{koalpaca2023}. Each triplet contains a Korean instruction, a fully Korean chosen response, and a synthetically generated code-mixed rejected response, created by injecting randomly selected English or Chinese tokens—translated via the Google Translate API—at random word-level positions.

To ensure high linguistic purity, we applied the following preprocessing steps: 
(1) filtered for chosen responses written entirely in Korean, guaranteeing a WPR and LPR of 1.0; 
(2) applied string normalization (e.g., whitespace trimming) to instruction, chosen, and rejected fields.

\textbf{OIG (Fully English Rejection):}
We constructed a triplet dataset using the OIG-small-chip2-ko\footnote{\url{https://huggingface.co/datasets/heegyu/OIG-small-chip2-ko}} corpus, which contains over 210K instruction-response pairs translated into Korean from the original English OIG dataset~\citep{oig2022}. 
Each triplet comprises a Korean instruction, a fully Korean chosen response, and a fully English rejected response. 
This dataset is designed to evaluate the model's ability to distinguish between clearly separated linguistic domains.

We applied several preprocessing steps to improve data quality: 
(1) applied string normalization; 
(2) filtered for chosen responses containing only Korean text; 
(3) discarded samples where the length ratio between chosen and rejected responses fell outside the range of 0.4 to 2.0; 
(4) removed duplicate instructions.
Each dataset contains approximately 10,000 instruction-response triplets, selected for linguistic consistency and diversity.

\textbf{HC3 (Fully English Rejection):}
We also constructed dataset using the HC3-ko\footnote{\url{https://huggingface.co/datasets/nayohan/HC3-ko}}, which contains 24.3k instruction pairs, each containing a human-written and a GPT-generated response, translated into Korean ~\citep{guo2023close, hc3ko}.


Each triplet contains a Korean instruction, a fully Korean chosen response, and a synthetically generated code-mixed rejected response.
This dataset is designed to evaluate the model's generalizing ability to use the unseen data during training.

We applied several preprocessing steps to improve data quality: 
(1) applied string normalization; 
(2) filtered for chosen responses containing only Korean text; 
(3) discarded samples where the length ratio between chosen and rejected responses fell outside the range of 0.4 to 2.0; 
(4) removed duplicate instructions.
(5) removed responses exhibiting generation failures caused by the language model, such as repeated phrases or malformed outputs due to server errors.


\section{ORPO Training Configuration}
\label{appendix:orpo-training-config}

Table~\ref{tab:orpo-config} outlines the training configuration used for ORPO fine-tuning. Both SmolLM2-1.7B and OLMo-2-1124-7B were trained for 3 epochs with a global batch size of 128. ORPO’s weighting coefficient $\beta$ was set to 0.1 across experiments, and training was performed using the DeepSpeed ZeRO-2 framework.

\begin{table}[h]
\centering
\fontsize{7}{9}\selectfont
\begin{tabular}{lcc}
\toprule
\textbf{Parameter} & \textbf{SmolLM2-1.7B (ORPO)} & \textbf{OLMo2-7B (ORPO)} \\
\midrule
GPUs & A6000 × 1 & H100 × 2 \\
Max sequence length & 8192 & 4096 \\
Micro batch size & 8 & 8 \\
Gradient accumulation & 16 & 8 \\
Global batch size & 128 & 128 \\
Training steps & 223 & 223 \\
Epochs & 3 & 3 \\
ORPO $\beta$ value & 0.1 & 0.1 \\
Optimizer & AdamW & AdamW \\
Framework & DeepSpeed ZeRO-2 & DeepSpeed ZeRO-2 \\
\bottomrule
\end{tabular}
\caption{\footnotesize Training configuration for ORPO fine-tuning on SmolLM2 and OLMo2 models.}
\label{tab:orpo-config}
\end{table}

\section{Average loss tracking for OLMo2 }
\label{appendix:loss-olmo}

\begin{figure}[h]
    \centering
    \includegraphics[width=1.0\columnwidth]{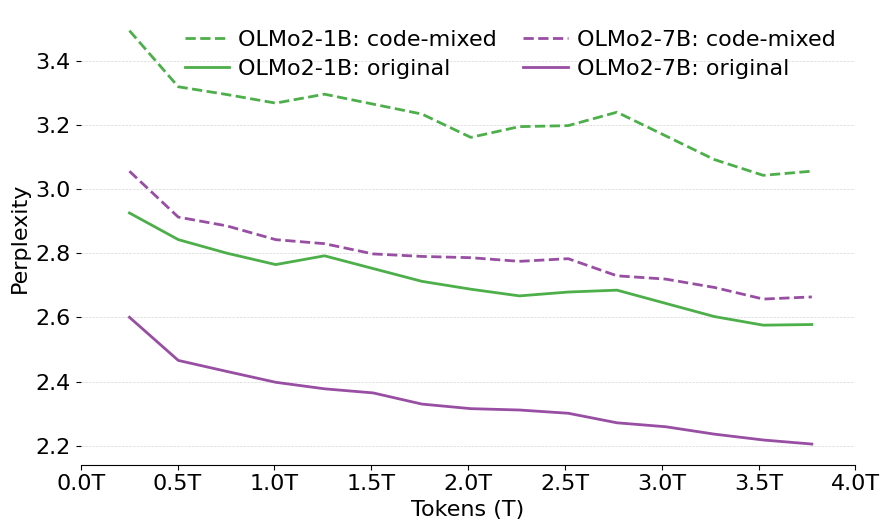}
    \caption{The average loss of original (monolingual) and code-mixed responses across training checkpoints for OLMo2 models.}
    \label{fig:loss-olmo}
\end{figure}

To assess whether the failure to penalize language confusion generalizes across architectures, we also tracked the loss trajectories of OLMo2 models (1B and 7B) throughout pretraining. 
As shown in Figure~\ref{fig:loss-olmo}, both original and code-mixed responses exhibit a steady decrease in loss, mirroring the trend observed in SmolLM2 (Figure~\ref{fig:loss_smollm}). 
Despite the increase in model capacity, the gap between two responses does not widen. This suggests that pretraining objectives alone may not induce meaningful linguistic preferences.

\section{Samples of different levels of WPR and LPR}
\label{appendix:different_levels_of_wpr_lpr}

\begin{figure*}[h]
    \centering
    \includegraphics[width=\textwidth]{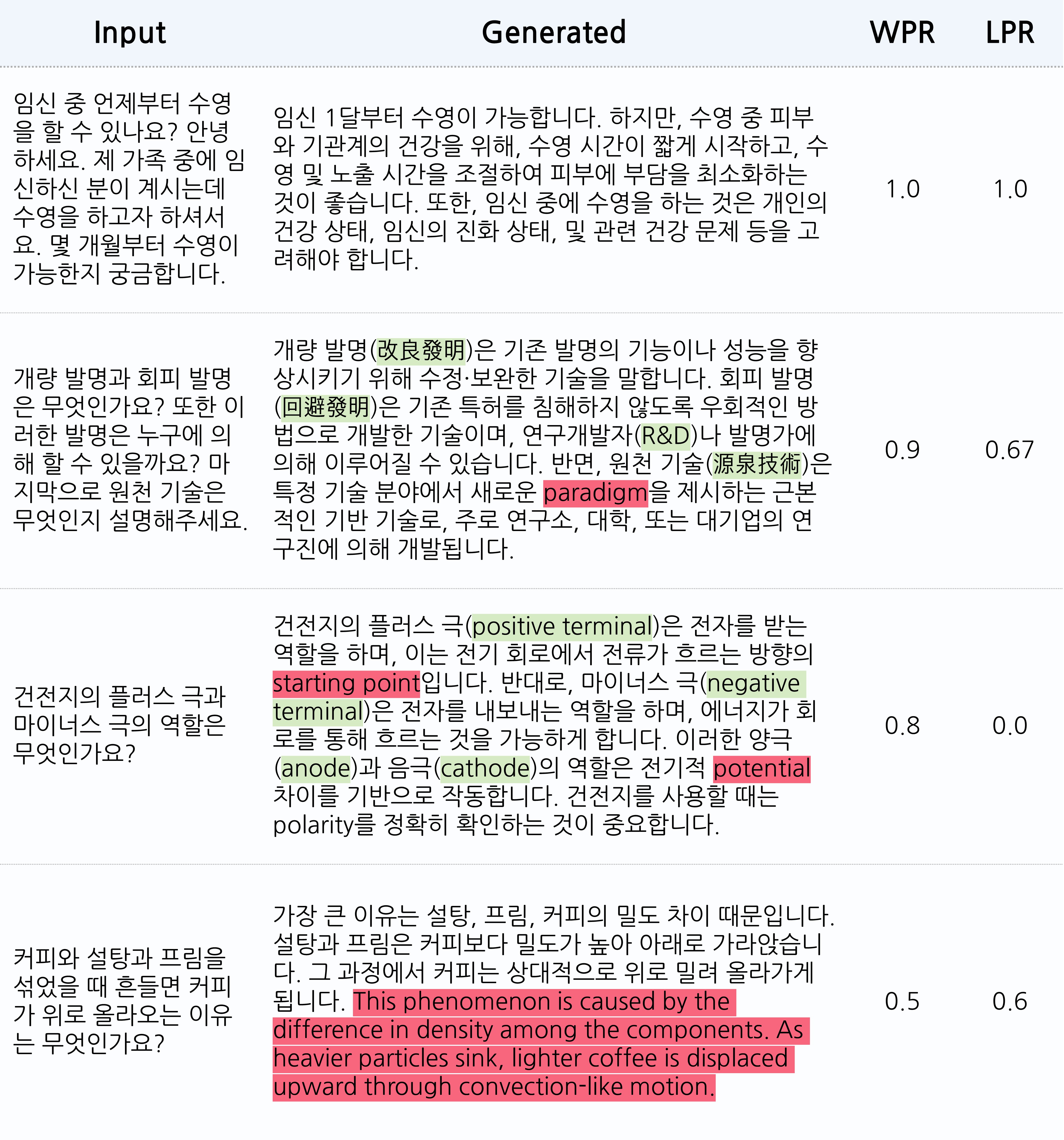}
    \caption{Samples of generated responses at varying WPR and LPR levels}
    \label{fig:wpr_lpr_example}
\end{figure*}

To enable interpretable comparisons across models, we report the proportion of generations that exceed a threshold of 0.9 for both WPR and LPR. 
This threshold was chosen based on manual inspection by a native Korean speaker, who reviewed a large number of generated samples and heuristically identified 0.9 as a practical cutoff that separates mostly monolingual responses from visibly code-mixed ones.
This level of tolerance allows minor lexical variation (e.g., loanwords, numerals) while still maintaining strong target-language alignment. 
It also aligns with real world expectations for language consistency, particularly in Korean, where partial foreign-language inclusions are not uncommon but still undesirable in many contexts.
Representative examples illustrating this thresholding effect are shown in Figure~\ref{fig:wpr_lpr_example}.

\section{Generation-level evaluation: other models }
\label{appendix: WPR/LPR results across 3 levels of temperatures}

In addition to ORPO, we evaluate two other fine-tuning methods: Supervised Fine-Tuning (SFT) and Direct Preference Optimization (DPO) across multiple decoding temperatures and model families (SmolLM2, OLMo2). 

Direct Preference Optimization (DPO) is a preference-based tuning method that trains models to maximize the log-probability margin between preferred and rejected responses~\citep{rafailov2023direct}.

Table~\ref{tab:dpo-setup} describes the detailed training configurations used for DPO fine-tuning.
All settings were selected to closely match the original DPO implementation where possible.

Table~\ref{appendix:smollm_sft_dpo_orpo} and Table~\ref{appendix:olmo_sft_dpo_orpo} summarize the generation performance of each model across three decoding temperatures (0.7, 1.0, 1.2) and three fine-tuning methods (SFT, DPO, ORPO). 
We report four key metrics: the ratio of outputs with WPR > 0.9, LPR > 0.9, average WPR, and average LPR.

Across both model families, ORPO consistently outperforms DPO and performs on par with or slightly better than SFT in terms of language fidelity. 
In particular, ORPO maintains near-perfect WPR and LPR values across all temperature settings, while DPO exhibits significant degradation at higher temperatures, most notably on the OLMo2 model at temperature 1.2 (LPR > 0.9 ratio drops to 52.1\%. SFT remains relatively stable across temperatures.


\begin{table}[h]
\centering
\fontsize{7}{9}\selectfont
\begin{tabular}{lcc}
\toprule
\textbf{Parameter} & \textbf{SmolLM2-1.7B (DPO)} & \textbf{OLMo2-7B (DPO)} \\
\midrule
GPUs & A6000 × 1 & A6000 × 4 \\
Dataset size & 10,000 & 10,000 \\
Max sequence length & 8192 & 4096 \\
Micro batch size & 8 & 4 \\
Gradient accumulation & 8 & 4 \\
Global batch size & 64 & 64 \\
Training steps & 467 & 467 \\
DPO $\beta$ value & 0.1 & 0.1 \\
Optimizer & RMSprop & RMSprop \\
Framework & DeepSpeed ZeRO-2 & DeepSpeed ZeRO-2 \\
\bottomrule
\end{tabular}
\caption{\footnotesize Training configuration for DPO fine-tuning on SmolLM2 and OLMo2 models.}
\label{tab:dpo-setup}
\end{table}

\begin{table*}[h]
\centering
\small
\caption{Performance of SmolLM2 across temperature and tuning methods (SFT, DPO, ORPO)}
\label{appendix:smollm_sft_dpo_orpo}
\begin{tabular}{lccc|ccc|ccc}
\toprule
\multirow{2}{*}{\textbf{Metric}} 
& \multicolumn{3}{c|}{\textbf{temperature = 0.7}} 
& \multicolumn{3}{c|}{\textbf{temperature = 1.0}} 
& \multicolumn{3}{c}{\textbf{temperature = 1.2}} \\
\cmidrule(lr){2-4} \cmidrule(lr){5-7} \cmidrule(lr){8-10}
&  \textbf{SFT} & \textbf{DPO} & \textbf{ORPO}
  & \textbf{SFT} & \textbf{DPO} & \textbf{ORPO}
  & \textbf{SFT} & \textbf{DPO} & \textbf{ORPO} \\
\midrule
WPR > 0.9 ratio &  99.9\% & 94.2\% & 100.0\% & 100.0\% & 96.9\% & 100.0\% & 100.0\% & 95.0\% & 100.0\% \\
LPR > 0.9 ratio & 99.8\% & 92.3\% & 99.9\% & 100.0\% & 94.4\% & 100.0\% & 99.7\% & 90.5\% & 99.9\% \\
Average WPR     & 0.9998 & 0.9760 & 0.9999 & 1.0000 & 0.9857 & 1.0000 & 0.9998 & 0.9823 & 0.9999 \\
Average LPR     &  0.9994 & 0.9705 & 0.9996 & 1.0000 & 0.9780 & 1.0000 & 0.9993 & 0.9629 & 0.9999 \\
\bottomrule
\end{tabular}
\end{table*}

\begin{table*}[h]
\centering
\small
\caption{Performance of OLMo2 across temperature and tuning methods (SFT, DPO, ORPO)}
\label{appendix:olmo_sft_dpo_orpo}
\begin{tabular}{lccc|ccc|ccc}
\toprule
\multirow{2}{*}{\textbf{Metric}} 
& \multicolumn{3}{c|}{\textbf{temperature = 0.7}} 
& \multicolumn{3}{c|}{\textbf{temperature = 1.0}} 
& \multicolumn{3}{c}{\textbf{temperature = 1.2}} \\
\cmidrule(lr){2-4} \cmidrule(lr){5-7} \cmidrule(lr){8-10}
& \textbf{SFT} & \textbf{DPO} & \textbf{ORPO}
  & \textbf{SFT} & \textbf{DPO} & \textbf{ORPO}
  & \textbf{SFT} & \textbf{DPO} & \textbf{ORPO} \\
\midrule
WPR > 0.9 ratio   & 99.8\% & 99.5\% & 99.8\% & 99.9\% & 99.4\% & 99.9\% & 99.1\% & 94.4\% & 99.0\% \\
LPR > 0.9 ratio   & 99.7\% & 92.7\% & 99.7\% & 99.8\% & 89.4\% & 99.8\% & 96.8\% & 52.1\% & 96.8\% \\
Average WPR       & 0.9996 & 0.9959 & 0.9998 & 0.9998 & 0.9938 & 0.9998 & 0.9970 & 0.9649 & 0.9962 \\
Average LPR       & 0.9988 & 0.9847 & 0.9992 & 0.9997 & 0.9791 & 0.9995 & 0.9915 & 0.8897 & 0.9881 \\
\bottomrule
\end{tabular}
\end{table*}

\end{document}